\documentclass{article} 
\usepackage{RobustML_iclr2021_conference,times}
\usepackage{hyperref}
\usepackage{url}
\title{Interpretable Mixture density estimation \\ by use of differentiable tree-module}
\author{Ryuichi Kanoh \\Mobility Technologies Co., Ltd. \\ \texttt{ryuichi.kano@mo-t.com} \And
Tomu Yanabe \\ Mobility Technologies Co., Ltd. \\ \texttt{tomu.yanabe@mo-t.com} 
}
\usepackage{graphicx}
\usepackage[ruled,vlined]{algorithm2e}
\usepackage{gensymb}
\usepackage{amsmath}
\usepackage{here}
\usepackage{wrapfig}
\usepackage{nicefrac}       
\usepackage{siunitx}
\iclrfinalcopy 
\begin{document}
\maketitle
\begin{abstract}
In order to develop reliable services using machine learning, it is important to understand the uncertainty of the model outputs. Often the probability distribution that the prediction target follows has a complex shape, and a mixture distribution is assumed as a distribution that uncertainty follows. Since the output of mixture density estimation is complicated, its interpretability becomes important when considering its use in real services. In this paper, we propose a method for mixture density estimation that utilizes an interpretable tree structure. Further, a fast inference procedure based on time-invariant information cache achieves both high speed and interpretability.
\end{abstract}
\section{Introduction}
Machine learning (ML) is becoming more and more popular, and is being used in various services. When the outputs of ML are used to determine the behavior of a service, probabilistic prediction \citep{pmlr-v48-gal16,NIPS2017_9ef2ed4b,pmlr-v119-duan20a} is attracting attention from the viewpoint of robustness. In the case of point prediction, the model's output is a single representative value. However, with the point prediction output, we cannot understand how confident the ML model is. Therefore, probabilistic prediction, in which the output is in the form of a probability distribution rather than a single value, has become an essential technique.
Note that the probability distribution of targets does not always have a simple form (e.g., single Gaussian). In the real world, the target distribution is often a mixture distribution (e.g., Gaussian mixture). Although several ML models that use mixture distributions as output have been proposed, there are still some issues in terms of interpretability. In particular, since the output results are more complicated than point prediction or probabilistic prediction with a single distribution, understanding how mixture distributions are formed in the ML models becomes an important topic.

In this paper, we propose an interpretable method for the mixture density estimation using a tree-structured module. In particular, using the proposed model makes it easy to capture how a specific distribution is selected among multiple candidate distributions. 

Assuming a real-world service, we consider a situation where the feature given as input to the ML model is a mixture of time-variant and time-invariant information\footnote{Assume future vehicle counts on each road-segment is a target of the prediction. In this example, weather, temperature, and the number of surrounding vehicles can be time-variant information. In contrast, road length, number of lanes on the road, and road location are time-invariant information.}. Under such a situation, we propose a method that makes part of the inference process pre-computable by efficiently handling time-invariant input. It allows us to achieve both high interpretability and high speed.

\section{Related work}
\subsection{Mixture density network} \label{subsec:mdn}
\begin{figure}
    \centering
    \includegraphics[width=12cm]{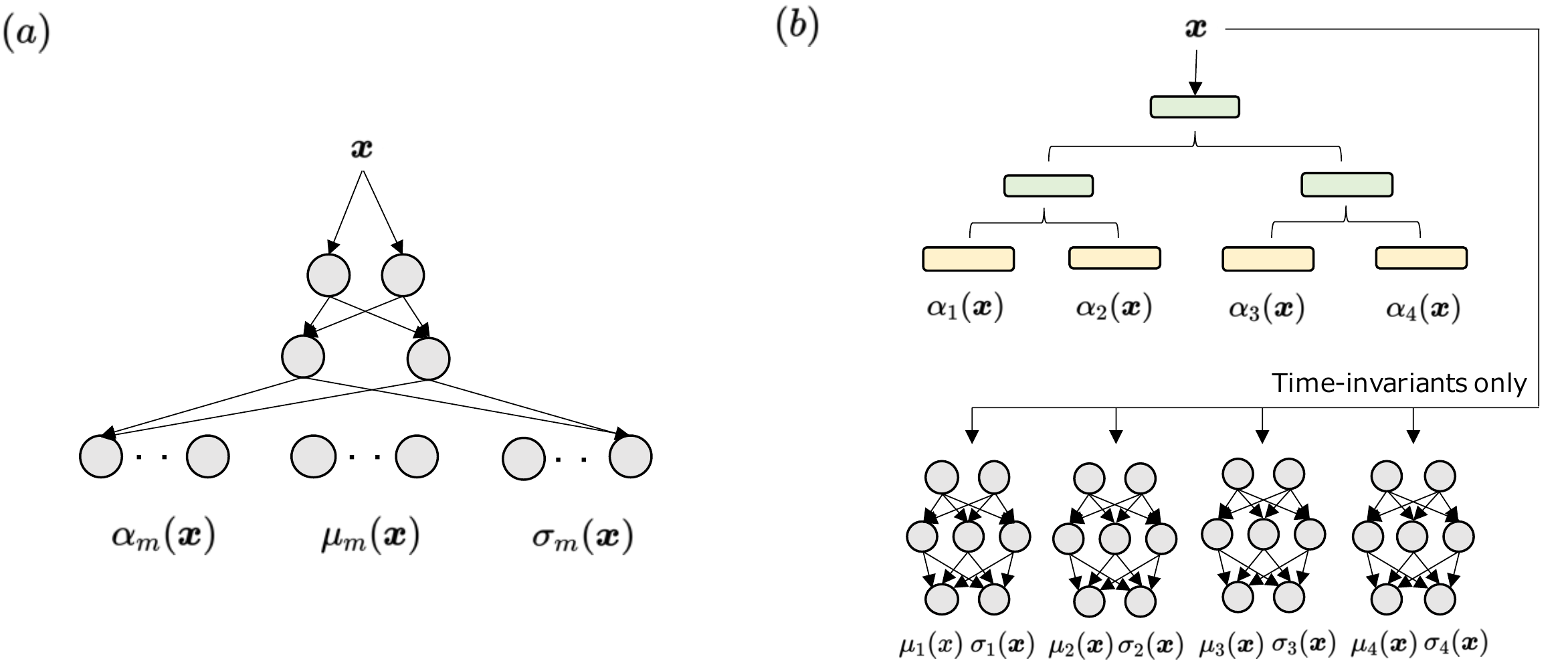}
    \caption{(a): Model architecture of Mixture Density Network. (b): Proposed model architecture.}
    \label{fig:mdn}
\end{figure}
\emph{Mixture density network} (MDN, \citet{Bishop94mixturedensity}) is an ML model that takes the output as formulated by a mixture of multiple distributions. Figure~\ref{fig:mdn}(a) shows an architecture of MDN. To represent a mixture distribution consisting of $M$ Gaussians $\sum_{m=1}^{M} \alpha_{m}(\boldsymbol{x})~\mathcal{N}\left(\mu_{m}(\boldsymbol{x}), \sigma_{m}(\boldsymbol{x})\right)$,
MDN outputs a collection of $M$ means $\mu_{m}(\boldsymbol{x})$, variances $\sigma_{m}(\boldsymbol{x})$ and weights $\alpha_{m}(\boldsymbol{x})$, where $\boldsymbol{x}$ is an input vector and $\mathcal{N}$ represents a Gaussian.
The MDN has a structure like multi-layer perceptron (MLP), and the hidden layers are shared to the output of $\mu_{m}(\boldsymbol{x})$, $\sigma_{m}(\boldsymbol{x})$, and $\alpha_{m}(\boldsymbol{x})$.
In practice, there are some constraints for parameters. First, weights must be normalized as $\sum_{i=1}^{m} \alpha_{m}(\boldsymbol{x})=1$. Therefore, the softmax function is applied for achieving constraints. Second, variances must not be negative. For this reason, the output of the model $\sigma_{m}(\boldsymbol{x})$ is exponentiated, and interpreted as variances. With obtained distribution parameters, negative log-likelihood (NLL) is minimized during the training.

Despite the simple structure, this MLP-based model does not the have high interpretability. Therefore, we propose a method that combines the benefits of the tree-structured model.
\subsection{Soft-Tree} \label{subsec:tree}
\begin{wrapfigure}{r}[10pt]{0.5\textwidth}
    \centering
    \includegraphics[width=7.0cm]{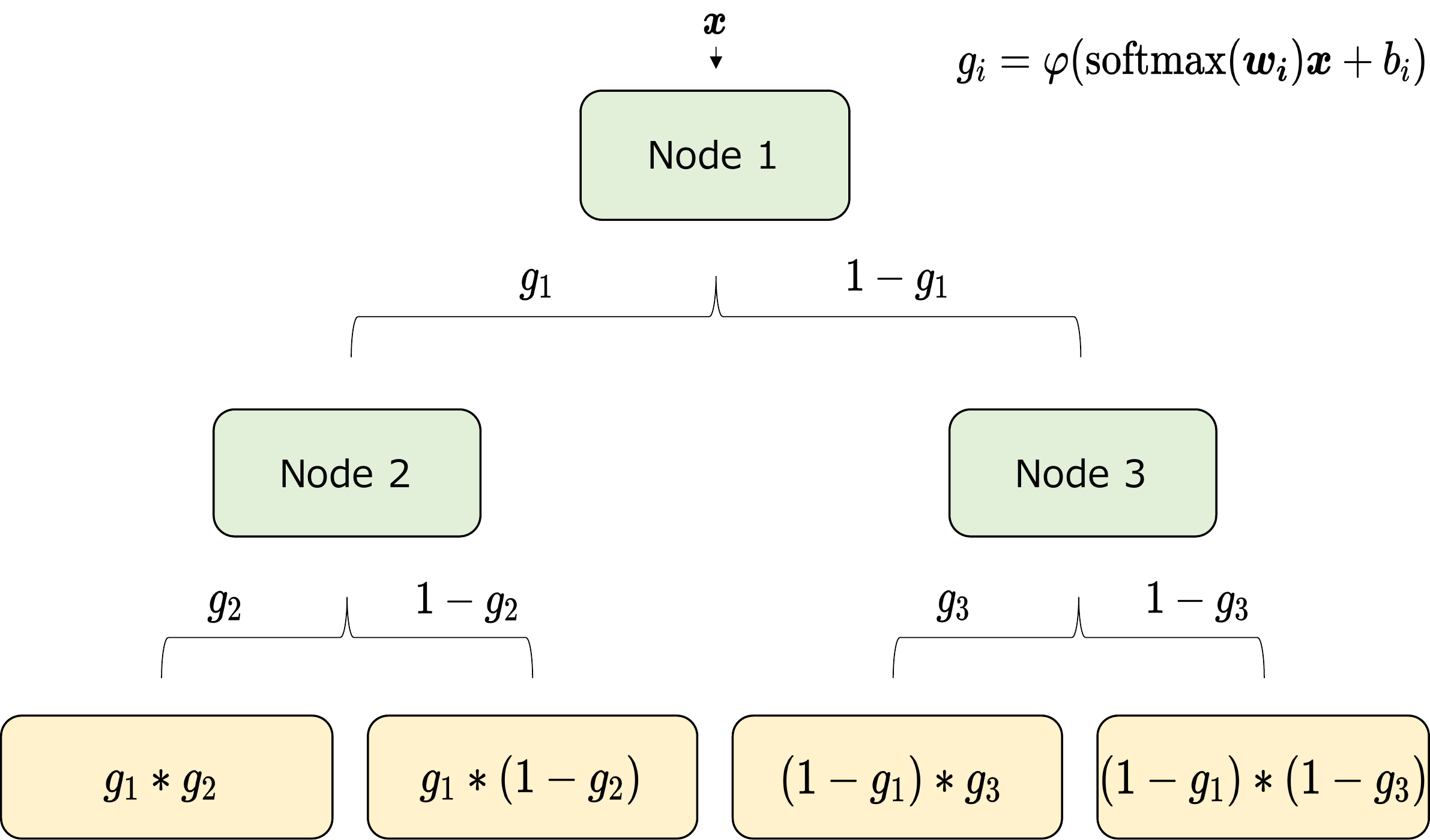}
    \caption{Soft splitting operation}
    \label{fig:tree_modules}
\end{wrapfigure}
A (hard) decision tree is an example of an ML model with high interpretability. However, since splitting search is a non-differentiable operation, it is not easy to combine differentiable models like a neural network. However, it is still possible to build a model that behaves like a decision tree using only differentiable operations \citep{DBLP:journals/corr/abs-1711-09784}. 

Figure~\ref{fig:tree_modules} shows a soft splitting operation. While hard decision trees decisively split data to either the right or the left, Soft-Trees splits data probabilistically. 
Each tree node $i$ has a trainable filter ${\boldsymbol{w_i}}$ and a bias ${b_i}$. Splitting is expressed using
\begin{equation}
    g_{i}=\varphi\left(\operatorname{softmax}\left(\boldsymbol{w}_{i}\right) \boldsymbol{x}+b_{i}\right), \label{eq:split}
\end{equation}
where $\varphi$ is an activation function like sigmoid that satisfies $0.0 \leq \varphi(x) \leq 1.0$, $\lim_{x \rightarrow \infty} \varphi (x) = 1.0$, and $ \lim_{x \rightarrow -\infty} \varphi (x)=0.0$.  In analogy with the hard decision tree, $\operatorname{softmax}\left(\boldsymbol{w}_{i}\right)$ corresponds to the process of feature selection used for tree splitting\footnote{To emphasize the analogy of feature selection, filter weight ${\boldsymbol{w_i}}$ is often passed through an activation function such as softmax or entmax \citep{entmax, Popov2020Neural}. Note that if $\operatorname{softmax}\left(\boldsymbol{w}_{i}\right)$ only contains $1.0$ and $0.0$, its role is equivalent to the hard feature selection.}, and ${b_i}$ corresponds to the process of threshold determination for splitting.
At each split, the data distribution split to the right and left node are
$r_r = r_{o} * g_i$ and $r_l = r_{o} * \left( 1- g_i\right)$, where splitting node has $r_{o}$ of data.
If $g_i$ is exactly $1.0$ or $0.0$, then the split process is equivalent to a hard decision tree. Performing this operation recursively, the leaves at the end of the tree-module have the data distribution corresponding to each leaf. This leaf distribution is used for weights on ensembling of the outputs of leaf-modules. 
The leaf-module can be a trainable constant scaler (that does not depend on $\boldsymbol{x}$) like a typical decision tree, or it can be another ML model like MLP.

Although such a tree-structured model has been proposed, how to perform mixture density estimation by using this model has not been studied well.

\section{Mixture density estimation with tree-module}
\subsection{Architecture}
Figure~\ref{fig:mdn}(b) shows a proposed model architecture. Compared with a simple MDN (Section~\ref{subsec:mdn}), there are two major differences. 
The first is the method for estimating $\alpha_{m}(\boldsymbol{x})$. While MDN outputs $\alpha_{m}(\boldsymbol{x})$ using the same procedure as $\mu_{m}(\boldsymbol{x})$ and $\sigma_{m}(\boldsymbol{x})$, the proposed method calculates $\alpha_{m}(\boldsymbol{x})$ using leaf data distribution of the Soft-Tree (Section~\ref{subsec:tree}). Because of this architecture, it is easy to capture how a particular distribution is selected among multiple candidate distributions. 
Second, while MDN shares the intermediate layers of a neural network and outputs multiple distribution parameters from it, in the proposed method, an MDN that outputs only one distribution is used for each leaf of the tree-module. This makes it easy to interpret what each leaf means. Since $M=1$ for each leaf MDN, there is no need to consider $\alpha_{m}(\boldsymbol{x})$. Note that only time-invariant features are used for inputs of leaf-modules. The reasons are described in the following (Section~\ref{subsec:speed}).
\subsection{Fast inference with pre-computation} \label{subsec:speed}
Since many neural networks (MDNs that outputs parameters of a single distribution) are used for each leaf-module, the number of parameters of the whole model is larger than simple MDN. Therefore, even with interpretability, processing speed is expected to be slow.
As a countermeasure, we use only time-invariant information as input of the leaf-modules. In that case, the leaf-module's output can be pre-computed, and only the tree-module needs to be calculated during the inference. Such a constraint achieves high-speed processing in real-world services where time-variant and time-invariant information are mixed.

Table~\ref{tbl:cost} shows a comparison of the number of parameters in the model between Soft-Tree and MLP. Since the number of elements of the mixture distribution is not expected to be very large when predicting a mixture distribution, the tree's depth is expected not to be large. Assuming that the number of hidden units in MLP $(W)$ is $100$, the number of input features is $30$, the number of output values is $2$, and the depth of MLP $(L)$ and Soft-Tree $(D)$ is $3$, then the number of parameters in Soft-Tree is $210$, and the number of parameters in MLP is $13,200$. From this example, we see that the amount of parameters in the tree-module required for actual inference is not large.
There are techniques to reduce the amount of calculation for Soft-Tree, that avoids exponential order computational complexity. In addition, we can speed up the training process by making splitting sparse like a hard decision tree. Details are in Appendix~\ref{app:tree_speed}.
\begin{table}[htb]
\centering
  \begin{tabular}{|c||c|} \hline
    model & \#~parameters \\ \hline
    Soft-Tree & $\mbox{len}(\mbox{input})(2^{D+1}-1)$ \\ 
    MLP (e.g., leaf-modules) & $W(\mbox{len}(\mbox{input})+\mbox{len}(\mbox{output}))+W^2 (L-2)$ \\ \hline
  \end{tabular}
  \caption{A number of parameters for $L$ layer MLP with $W$ hidden units per each layer, and Soft-Tree with $D$ depth. The Soft-Tree is assumed to be a complete binary tree. For simplicity, the number of bias terms is not counted. We assume $W$ does not change for each layer.}
  \label{tbl:cost}
\end{table}
\section{Numerical experiments}
\subsection{Setup}
We use real-world data obtained by MOV\footnote{https://m-o-v.jp/} (renamed as GO\footnote{https://go.mo-t.com/} in September 2020), a Japanese taxi ride-hailing service by Mobility Technologies\footnote{https://mo-t.com/}. In this experiment, the time duration between entering and exiting the road-segments of taxi stands is the prediction target. We made a prediction for each record that occurs each time a vehicle passes by. Since the target does not take negative values and is represented by a superposition of lognormal distributions, the logarithm is taken to the prediction target. Other details of experiments are in Appendix~\ref{app:setup}.
\subsection{Result}
\begin{figure}
    \centering
    \includegraphics[width=14cm]{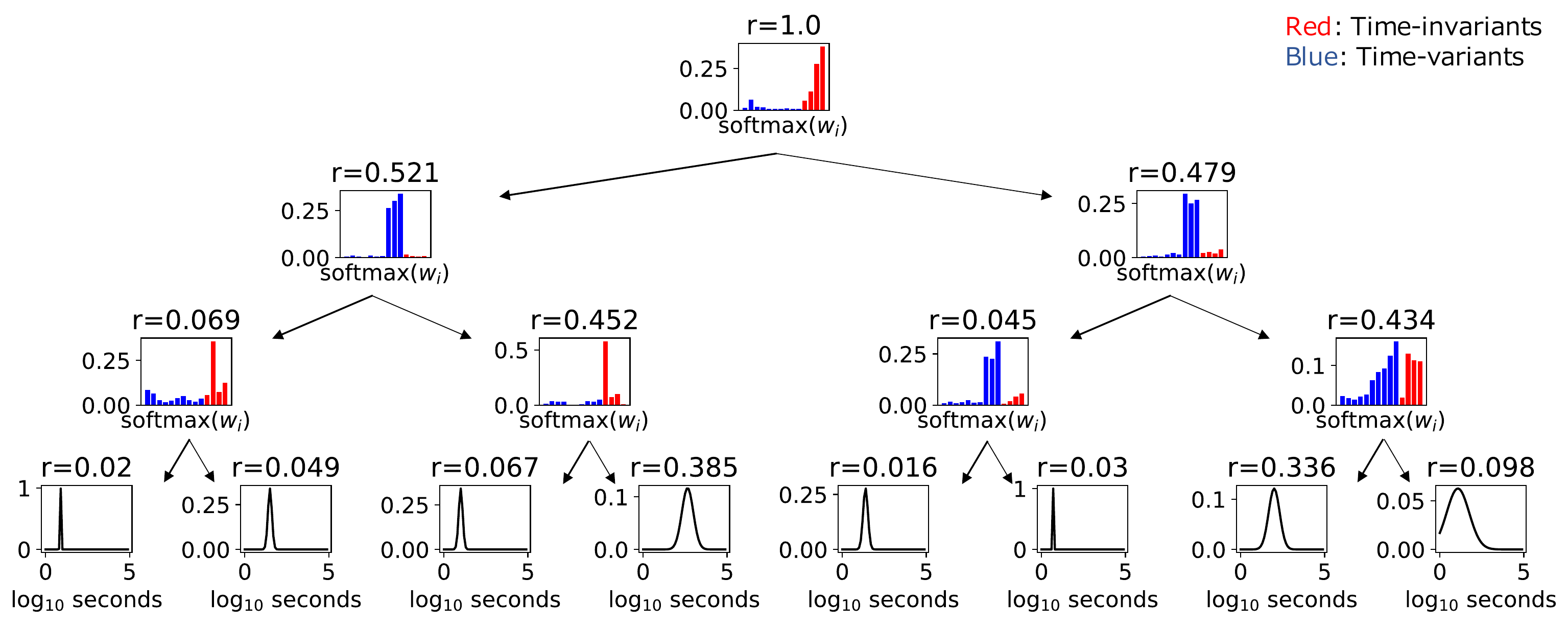}
    \caption{Trained proposed model. For a tree-module, the length of each bar is the softmax$(\boldsymbol{w_{i}})$. The red and blue colors in the bar plots correspond to softmax$(\boldsymbol{w_{i}})$ for time-invariant and time-variant features, respectively.
    For leaf-modules, their output probability distribution for a single input feature set is plotted. 
    At the top of each plot, how much percentage is flowing to that node is stated. The softmax$(\boldsymbol{w_{i}})$ does not change when the input data changes. In contrast, the leaves' information, such as the probability distribution and how much data is distributed to each node, changes with input data.}
    \label{fig:visualization}
\end{figure}

\begin{table}[htb]
\centering
  \begin{tabular}{|c|c||c|} \hline
    & model & NLL (lower is better) \\ \hline
    (1) & MDN & $-0.1937 \pm 0.2144$  \\
    (2) & Soft-Tree with constant leaves & $0.9926 \pm 0.1148$ \\
    (3) & Proposed & $-0.4512 \pm 0.1925$ \\ 
    \hline
  \end{tabular}
  \caption{Comparison of probabilistic regression performance as measured by NLL. The mean and standard deviation of the NLL are summarized from experiments with $10$ different random seeds.}
  \label{tbl:exp}
\end{table}

Table~\ref{tbl:exp} shows a performance comparison. As benchmarks, we used (1) MDN shown in Figure~\ref{fig:mdn}(a), and (2) Soft-Tree with constant leaves that do not depend on $\boldsymbol{x}$. As for (2), the difference from the proposed method is that the output of the leaf-modules ($\mu_{m}$ and $\sigma_{m}$) do not depend on any input $\boldsymbol{x}$ during the inference, and therefore the candidates of the element of mixture distributions are common for all prediction results. Note that in the proposed method, the output of the leaf-modules depends on the time-invariant input features.
On the one side, compared to (1), the performance of the proposed method is better, even though the number of the parameters required for inference is smaller.
On the other hand, compared to (2), we can see the importance of changing the candidate elements of the mixture distribution for each input. Note that the computational complexity of (2) and the proposed method remains the same in inference due to the pre-computation.

Figure~\ref{fig:visualization} shows a trained proposed model. As shown in the figure, it is possible to visually check which features are responded to and how the data branched out and fell on the leaves. Each node of the tree-module has different features to focus on, indicating that each node has a different role. Since the leaf-modules that outputs a single Gaussian take only time-invariant features as input, the Gaussian visualized in the leaf-module does not change even if the time-variant features change. It is beneficial from the point of view of speed (Section~\ref{subsec:speed}) and interpretability. Since the role of each leaf does not change with time-variant input, it makes it easy to interpret when making comparisons between prediction outputs with the same time-invariant features (e.g., prediction outputs for the same road-segment at different times).

\bibliography{iclr2021_conference}
\bibliographystyle{iclr2021_conference}
\appendix
\section{Additional speedup for tree-module} \label{app:tree_speed}
\paragraph{Sparse Split}
A typical decision tree is a binary tree, where $100\%$ of the data is assigned to either the right node or the left node at the time of splitting. Consider we have a complete binary hard decision tree that has a depth of $4$. Although it has $15$ nodes in total, the split operation is not required for all nodes because of this sparszity.
Soft-Tree splitting does not always flow completely to either node, but there are a number of studies for achieving sparse splitting. For example, modifying the activation function $\varphi$ used in the splitting~(Equation~\eqref{eq:split} for taking the value of $0.0$ or $1.0$ strictly is one strategy. Notably, sparsity can achieve not only faster inference, but also faster training. \citep{ShazeerMMDLHD17, pmlr-v119-hazimeh20a, lepikhin2021gshard}. If the data distribution assigned to a leaf is exactly zero, the leaf-module calculation becomes unnecessary, and a large computational savings can be achieved.
\paragraph{Oblivious Tree}
An oblivious decision tree \citep{Popov2020Neural, NEURIPS2018_14491b75} uses the same splitting criterion across an entire level of the tree.
Because of its structure, the number of split process in the Soft-Tree can be reduced to the decision process for the depth of the tree.
\paragraph{Architecture Search}
In a general decision tree, the structure is not determined before training, and the tree grows during the training. As an analogy, \cite{pmlr-v97-tanno19a} proposed a method for growing Soft-Tree during training. There is a possibility that it allows us to use a smaller tree structure than the one we had pre-determined.
\section{Details of numerical experiments} \label{app:setup}
\begin{figure}
    \centering
    \includegraphics[width=14cm]{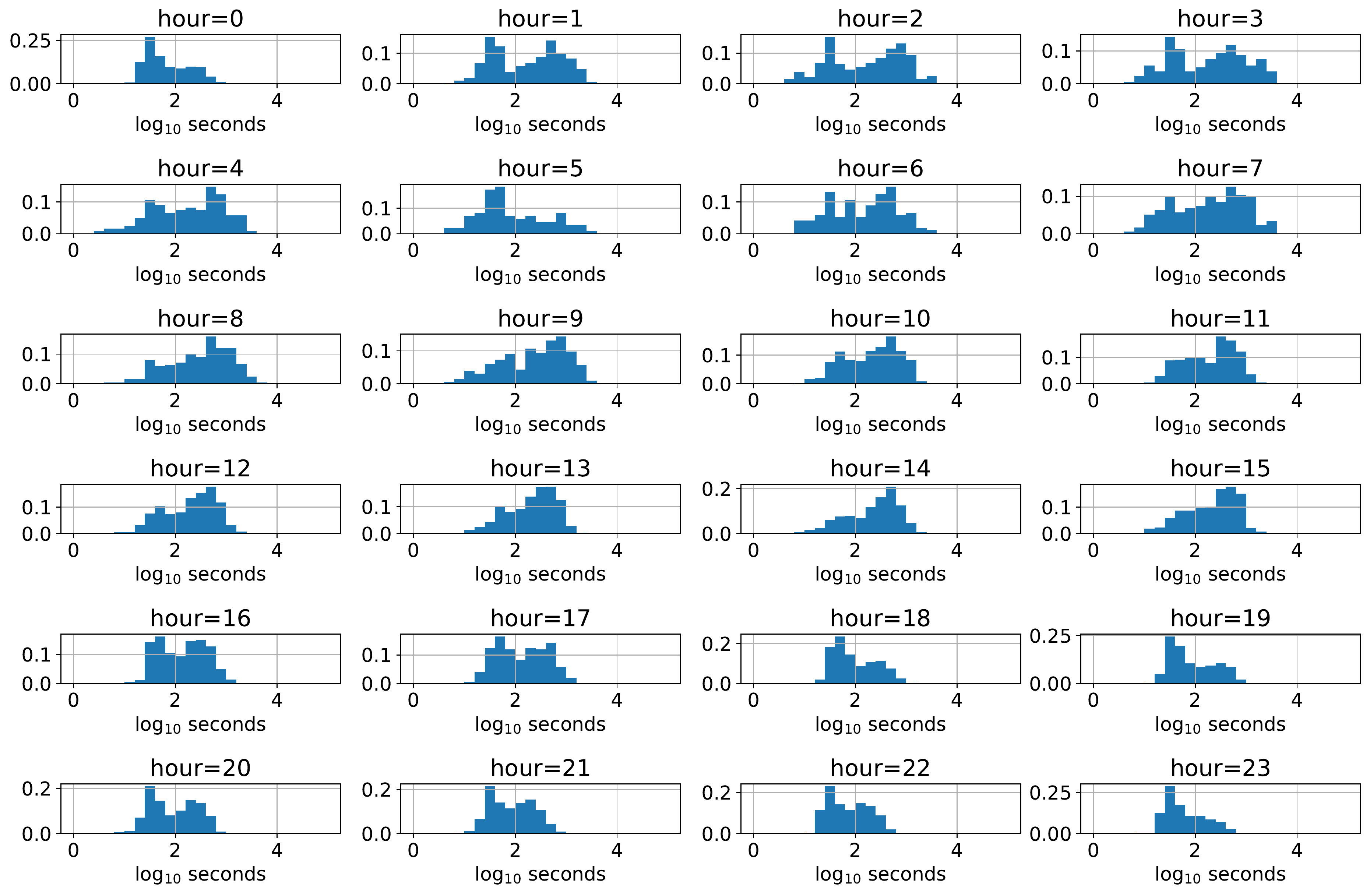}
    \caption{Distribution of the prediction target by an hour in a day in a road-segment}
    \label{fig:temporal-change}
\end{figure}

\paragraph{Dataset}
Vehicle position (GPS latitude, longitude), and their status (Looking for passengers, carrying passengers, heading to reservation location) are obtained every 3 seconds for each taxi. Hidden Markov map-matching \citep{newson2009hidden} is used for all GPS points. For predicting the time duration between entering and exiting the road-segments of  taxi stands, we use taxi stands that exist within a square with a side length of $10 \si{km}$, centered at latitude $=35.4583$ and longitude $=139.5625$, which vehicles passed through more than $10,000$ times in a month. In total, $939,954$ records are used for numerical experiments. Figure~\ref{fig:temporal-change} shows an example of the distribution of the prediction target by an hour in a day in a road-segment.
\paragraph{Training procedure}
Numerical experiments are conducted with the data acquired in January 2020. Datasets obtained on January $1$-$14$, $14$-$21$, and $21$-$28$ are used as train, valid, and test dataset, respectively. The performance is evaluated using the test data, and the values are reported when the training had the best performance for the valid data. We perform $2.0 \times 10^3$ optimization steps for minimizing the NLL. The prediction target is log-scaled. Batch size and learning rate are $2048$ and $1.0 \times 10^{-1}$, respectively. We use Adam optimizer \citep{DBLP:journals/corr/KingmaB14} with a cosine-annealing learning rate scheduler.
\paragraph{Model Architecture}
We use a Soft-Tree with a depth $D=3$ and an MLP with depth $L=2$ and width $W=50$ as the leaf-modules. Since the depth of the Soft-Tree is $3$, it can represent a mixture distribution consisting of at most $2^3=8$. In order to align the conditions, the benchmark MDN (Table~\ref{tbl:exp}) also outputs parameters of $8$ Gaussians. $W$ and $L$ are the same as benchmark MDN and proposed method.
\paragraph{Input features}
$14$ features are used for training and inference. There are $10$ time-variant features and $4$ time-invariant features. 
\begin{itemize}
    \item Time-variant features
    \begin{enumerate}
        \item Sine/Cosine transformed hour of a day
        \item Sine/Cosine transformed day of week
        \item Passed vehicle counts in recent $15$min/$30$min/$60$min
        \item Average time duration for passing the taxi stands in recent $15$min/$30$min/$60$min
    \end{enumerate}
    \item Time-invariant features
        \begin{enumerate}
            \item Latitude and longitude of the taxi stand
            \item Length of the road-segment of the taxi stand
            \item Total vehicle counts per taxi stand
        \end{enumerate}
\end{itemize}
All features are numerical. Standard scaling is applied for each feature to achieve zero-mean and unit-variance.
\end{document}